\documentclass[final]{cvpr}

\usepackage{times}
\usepackage{epsfig}
\usepackage{graphicx}
\usepackage{amsmath}
\usepackage{amssymb}

\usepackage{booktabs}
\usepackage{colortbl}
\usepackage[linesnumbered,ruled,vlined]{algorithm2e}

\usepackage{makecell}
\usepackage{multirow}
\usepackage{enumerate}
\newcommand{\PreserveBackslash}[1]{\let\temp=\\#1\let\\=\temp}
\newcolumntype{C}[1]{>{\PreserveBackslash\centering}p{#1}}
\newcolumntype{R}[1]{>{\PreserveBackslash\raggedleft}p{#1}}
\newcolumntype{L}[1]{>{\PreserveBackslash\raggedright}p{#1}}
\usepackage[pagebackref=true,breaklinks=true,letterpaper=true,colorlinks,bookmarks=false]{hyperref}

\newcommand{\tabincell}[2]{\begin{tabular}{@{}#1@{}}#2\end{tabular}}

\DeclareMathOperator*{\argmax}{arg\,max}
\newcommand{\BEST}[1]{\textbf{\textcolor[rgb]{1.00,0.00,0.00}{#1}}}
\newcommand{\SBEST}[1]{\textbf{\textcolor[rgb]{0.00,0.80,0.00}{#1}}}
\newcommand{\TBEST}[1]{\textbf{\textcolor[rgb]{0.00,0.00,1.00}{#1}}}

\def\cvprPaperID{7297} 
\def\confYear{CVPR 2021}

\begin{document}

\title{CRACT: Cascaded Regression-Align-Classification for Robust Visual Tracking}

\author{Heng Fan  \;\;\;\;\; Haibin Ling\\
Department of Computer Science, Stony Brook University, Stony Brook, NY USA\\
{\tt\small \{hefan,hling\}@cs.stonybrook.edu}
}

\maketitle

\begin{abstract}
   
   High quality object proposals are crucial in visual tracking algorithms that utilize region proposal network (RPN). Refinement of these proposals, typically by box regression and classification in parallel, has been popularly adopted to boost tracking performance. However, it still meets problems when dealing with complex and dynamic background. Thus motivated, in this paper we introduce an improved proposal refinement module, \textbf{Cascaded Regression-Align-Classification} (CRAC), which yields new state-of-the-art performances on many benchmarks. 
   
   First, having observed that the offsets from box regression can serve as guidance for proposal feature refinement, we design CRAC as a cascade of box regression, feature alignment and box classification. The key is to bridge box regression and classification via an alignment step, which leads to more accurate features for proposal classification with improved robustness. To address the variation in object appearance, we introduce an \textbf{identification-discrimination} component for box classification, which leverages offline reliable fine-grained template and online rich background information to distinguish the target from background. Moreover, we present \textbf{pyramid RoIAlign} that benefits CRAC by exploiting both the local and global cues of proposals. During inference, tracking proceeds by ranking all refined proposals and selecting the best one. In experiments on \textbf{seven} benchmarks including OTB-2015, UAV123, NfS, VOT-2018, TrackingNet, GOT-10k and LaSOT, our CRACT exhibits very promising results in comparison with state-of-the-art competitors and runs in real-time.
   
\end{abstract}

\section{Introduction}

\begin{figure}
	\centering
	\includegraphics[width=\linewidth]{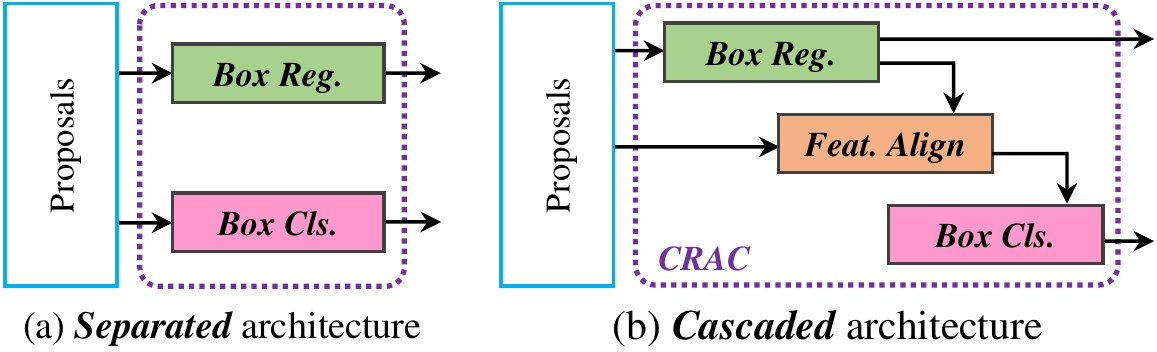}
	\caption{Different proposal refinement structures: separated box regression and classification in parallel (\eg,~\cite{wang2019spm,fan2019siamese}) in image (a) and our cascaded regression-align-classification (CRAC) in image (b). \emph{Best viewed in color and by zooming in}.}
	\label{fig:fig1}
\end{figure}

As one of the important problems in computer vision, visual tracking has many applications including video surveillance, intelligent vehicles, human-machine interaction, \etc\ Despite considerable progress made in recent years, robust tracking remains challenging because of many factors such as occlusion, distractor, scale changes, deformation, motion blur and so on~\cite{fan2019tracklinic}.

In this paper we focus on model-free single object tracking. Specifically, given the target in initial frame, a tracker aims at locating it in all subsequent frames by determining its position and scale. Inspired by the Siamese tracking algorithm~\cite{bertinetto2016fully} and the region proposal network (RPN)~\cite{ren2015faster}, SiamRPN~\cite{li2018high,li2019siamrpn++} formulates tracking as an one-shot inference problem and has attracted great attention owing to its excellent performance in both accuracy and speed. It simultaneously predicts classification results and regression offsets for a set of pre-defined anchors to generate proposals. Encouraged by the success of SiamRPN, improvement has been proposed (\eg,~\cite{wang2019spm,fan2019siamese}) with an additional refinement process, which further regresses and classifies \emph{in parallel} each proposal (see Figure~\ref{fig:fig1}(a)). Particularly, regression is used to adjust the locations and sizes of proposals for better accuracy, and classification to distinguish the target object from background in proposals for better robustness. 

Despite improvements achieved, trackers with the above proposal refinement still fail in presence of complex background because of degenerated classification, caused by two problems: (1) In classification task, the features of proposals are directly extracted based on their locations. The inaccuracy in these locations (\eg, due to large scale changes) may contaminate the proposal features (\eg, due to irrelevant background information) and consequently degrades classification results. (2) Background appearance information, which may vary over time and plays a crucial role in distinguishing target from similar objects, is ignored in classification and may hence cause drift to distractors in background. 

\subsection{Contribution}

Motivated by aforementioned observations, in this paper we design a new proposal refinement module to improve the robustness of visual tracking.

First, we introduce a novel simple yet effective cascade of regression-align-classification (CRAC) for proposal refinement, which is different than the parallel regression and classification utilized in existing approaches (Figure~\ref{fig:fig1} (a)). This design is motivated by the fact that the offsets from box regression can serve as guidance to sample more accurate proposal features. CRAC consists of three sequential steps, \ie, box regression, feature alignment and box classification, as shown in Figure~\ref{fig:fig1} (b). Specifically, {\it box regression} aims at further adjusting scales of proposals for better accuracy; {\it feature alignment} leverages offsets from box regression to better align proposals for improving feature quality; and {\it box classification} produces refined classification scores for aligned proposals. The key design in CRAC is to connect box regression and classification via an alignment step, instead of separating these two tasks. Such design enables more accurate features of aligned proposals, improving robustness of classification in refinement.

Then, to improve the robustness against background distractors, we develop an identification-discrimination component in the box classification step of CRAC. Specifically, the identifier learns \emph{offline} a distance measurement and utilizes reliable fine-grained target template to select the proposal most similar to target. The discriminator, drawing the inspiration from success in discriminative regression tracking~\cite{danelljan2019atom,danelljan2017eco,lu2018deep}, learns \emph{online} a discrete-sampling-based classification model using background and temporal appearance information to suppress similar objects in the proposals. By collaboration of identifier and discriminator, CRAC effectively inhibits distractors in the box classification step. 

Furthermore, to enhance representation of proposals, we introduce a pyramid RoIAlign (PRoIAlign) module for proposal feature extraction. PRoIAlign is capable of exploiting both local and global cues of proposals, and hence allows CRAC to deal with target deformation and rotation.

We integrate CRAC in the Siamese tracking framework to develop a new tracking algorithm named CRACT (\underline{CRAC} \underline{T}racker). CRACT first extracts a few coarse proposals via a Siamese-style network and then refines each proposal using CRAC.	Then the proposal with the highest classification score is selected to be target. In thorough experiments on seven benchmarks including OTB-2015~\cite{WuLY15}, UAV123~\cite{mueller2016benchmark}, NfS~\cite{kiani2017need}, VOT-2018~\cite{kristan2018sixth}, TrackingNet~\cite{muller2018trackingnet}, GOT-10k~\cite{huang2019got} and LaSOT~\cite{fan2019lasot}, our CRACT achieves new state-of-the-art results and significantly outperforms its Siamese baselines, while running in real-time. The implementation and results will be released upon publication of this work.

In summary, we make the following contributions.
\vspace{-0.5em}
\begin{enumerate}[1)]
	\setlength{\itemsep}{0pt}
	\setlength{\parsep}{0pt}
	\setlength{\parskip}{0pt}
	
	\item \emph{A new cascaded regression-align-classification (CRAC) module is developed for proposal refinement to improve the accuracy and robustness in tracking.}
	
	\item \emph{A novel identification-discrimination component is introduced to leverage offline and online learning of target and background information for handling distractors.}
	
	\item \emph{A pyramid RoIAlign strategy is designed to exploit both local and global cues of proposals for further improving robustness of CRAC.}
	
	\item \emph{A new tracker dubbed CRACT is developed based on the CRAC module, and achieves new state of-the-art results on numerous benchmarks.}
\end{enumerate}

\begin{figure*}
	\centering
	\includegraphics[width=\linewidth]{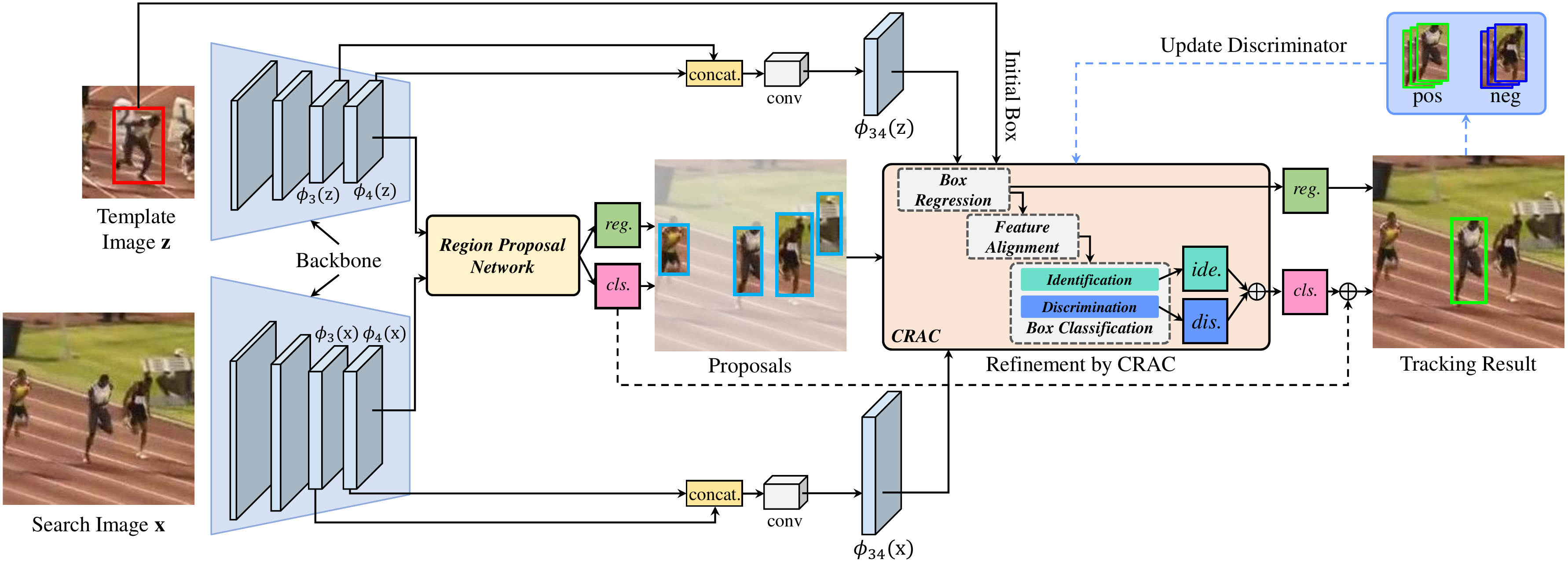}
	\caption{ Illustration of CRACT which first extracts a few coarse proposals (described in section~\ref{PRO}) and then refines each proposal with our cascaded RAC module (described in section~\ref{RAC}). The best proposal is selected based on coarse and refined classification scores to be tracking result. \emph{Best viewed in color and by zooming in}.}
	\label{fig:fig2}
\end{figure*}

\section{Related Work}

Visual object tracking has been extensively researched in recent decades. In this section, we discuss the most relevant work and refer readers to~\cite{smeulders2013visual,li2013survey,li2018deep,marvasti2019deep} for comprehensive surveys.

\vspace{0.5em}
\noindent
{\bf Siamese Tracking.} Treating tracking as searching for a region most similar to the initial target template, Siamese network has attracted great attention in tracking. The approach of~\cite{tao2016siamese} utilizes a Siamese network to learn a matching function from videos, and then uses it to search for the target object. Despite promising result, this approach runs slowly due to heavy computation. The work of~\cite{bertinetto2016fully} proposes a fully convolutional Siamese network (SiamFC) which efficiently computes the similarity scores of candidate regions. Owing to balanced accuracy and speed, SiamFC has been improved in many follow-ups~\cite{zhang2018structured,wang2018learning,he2018twofold,zhang2019deeper,guo2017learning,li2018high,li2019siamrpn++}. Among them, the work of~\cite{li2018high} introduces the SiamRPN by combining Siamese network and region proposal network~\cite{ren2015faster} for tracking, achieving more accurate results with faster speed. To improve SiamRPN in dealing with distractors, the work of~\cite{zhu2018distractor} leverages more negative training samples for learning a distractor-aware classifier. The approaches of~\cite{wang2019spm,fan2019siamese} cascade multiple stages to gradually improve the discrimination power of classification. In addition, for more accurate result, the approaches of~\cite{wang2019fast,yu2020deformable} integrate an additional segmentation branch into SiamRPN. More recently, anchor-free Siamese trackers~\cite{zhang2020ocean,chen2020siamese,guo2020siamcar} are proposed by predicting object bounding box offsets from a single pixel.

\vspace{0.5em}
\noindent
{\bf Cascade Structure in Tracking.} Cascade architecture has been a popular framework for vision tasks, and our CRACT also shares this idea for tracking. The work of~\cite{hua2015online} regards tracking as a proposal selection task and introduces a two-step tracker in which object proposals are first extracted and then classified with an online model. The approach of~\cite{wang2019spm}, based
on SiamRPN~\cite{li2018high}, presents a two-stage framework in which the proposals generated in the first stage are further identified and refined to choose the best one as the tracking result. The algorithm in~\cite{fan2019siamese} suggests a multi-stage framework that cascades multiple RPNs to improve performance of Siamese tracking.

\vspace{0.5em}
\noindent
{\bf Discriminative Regression Tracking.} Visual tracking with discriminative regression has demonstrated remarkable success recently. Among the most representative examples are correlation filter trackers~\cite{bolme2010visual,henriques2014high,kiani2015correlation} that formulate tracking as a rigid regression problem. Because of fast solution using fast Fourier transformation, this type of trackers usually run fast. Recently, motivated by powerful representation, deep feature has been applied in discriminative regression tracking~\cite{ma2015hierarchical,danelljan2017eco,dai2019visual}, significantly boosting performance. To further exploit the advantages of deep features, existing methods~\cite{song2017crest,lu2018deep,danelljan2019atom,bhat2019learning} propose to learn a convolutional regression model with the deep learning framework, which effectively improves performance. Notably, the
work of~\cite{danelljan2019atom} introduces a novel scale estimation approach
by IoU-Net~\cite{jiang2018acquisition}, leading to more accurate result.

\vspace{0.5em}
\noindent
{\bf Our Approach.} In this paper, we regard tracking as a proposal selection task. Our approach is related to but different from SiamRPN~\cite{li2018high} which treats tracking as one-shot proposal selection and may suffer from large scale changes and distractors. In contrast, we propose a novel CRAC refinement module to improve proposal selection and achieve better performance. Our method is also relevant to~\cite{wang2019spm,fan2019siamese} by sharing similar idea of refining proposals. However, unlike in~\cite{wang2019spm,fan2019siamese} that separately performs regression and classification for refinement, our method takes a cascade structure for refinement. Furthermore, different from~\cite{wang2019spm} using only local cues for proposal, we present pyramid RoIAlign to enhance proposals with both local and global information.

\section{Tracking with Cascaded Regression-Align-Classification}  

In this section, we formulate object tracking as selecting the best proposal and introduce a novel simple yet effective cascaded regression-align-classification (CRAC) module to refine proposals for such purpose.

As shown in Figure~\ref{fig:fig2}, our method contains proposal extraction and proposal refinement. In specific, we first use a Siamese region proposal network to filter out most low confident regions and keep only a few initial proposals. Then, each proposal is fed to the CRAC module for refinement of scale and classification results. During tracking, we rank all refined proposals using the initial and refined classification results, and the proposal with highest score is selected to be the final target. To maintain strong discriminative ability of our tracker, the discriminator in box classification of CRAC is online updated using intermediate results.

\subsection{Proposal Extraction}
\label{PRO}

The goal of proposal extraction is to filter out most negative candidates and retain a few initial proposals similar to target object. This procedure is crucial as one of the proposals from this stage determines the final tracking result. Therefore, it is required to be robust enough to include targets of interest into proposals and to avoid contamination from background. In addition, high efficiency is desired in the proposal extraction. Taking the above reasons into consideration, we leverage Siamese region proposal network, as in~\cite{li2018high,li2019siamrpn++,wang2019spm,fan2019siamese}, for proposal extraction.

The architecture of Siamese RPN contains two branches for target template $\mathbf{z}$ and search region $\mathbf{x}$, respectively. As illustrated in Figure~\ref{fig:fig2}, using ResNet~\cite{he2016deep} as backbone, we first extract the features $\phi_{4}(\mathbf{z})$ and $\phi_{4}(\mathbf{x})$ after block 4 for $\mathbf{z}$ and $\mathbf{x}$. Notice that, the feature extraction backbones for $\mathbf{z}$ and $\mathbf{x}$ share the same parameters. Then, $\phi_{4}(\mathbf{z})$ and $\phi_{4}(\mathbf{x})$ are fed to RPN, which simultaneously performs classification and regression for predefined anchors on search region (Please see architecture of RPN in the supplementary material). With the classification scores and regression offsets of anchors, we generate $N$ proposals using Non-maximum Suppression (NMS). We represent $N$ proposals as $\{p_{i}\}_{i=1}^{N}$, and classification result of $p_{i}$ is denoted as $c_{{i}}$. The loss $\ell_{\mathrm{rpn}}$ to train Siamese RPN comprises two parts including a cross entropy loss for classification and a smooth $L_1$ loss~\cite{girshick2015fast} for regression. We refer readers to~\cite{li2018high,girshick2015fast} for more details.

\begin{figure}
	\centering
	\includegraphics[width=\linewidth]{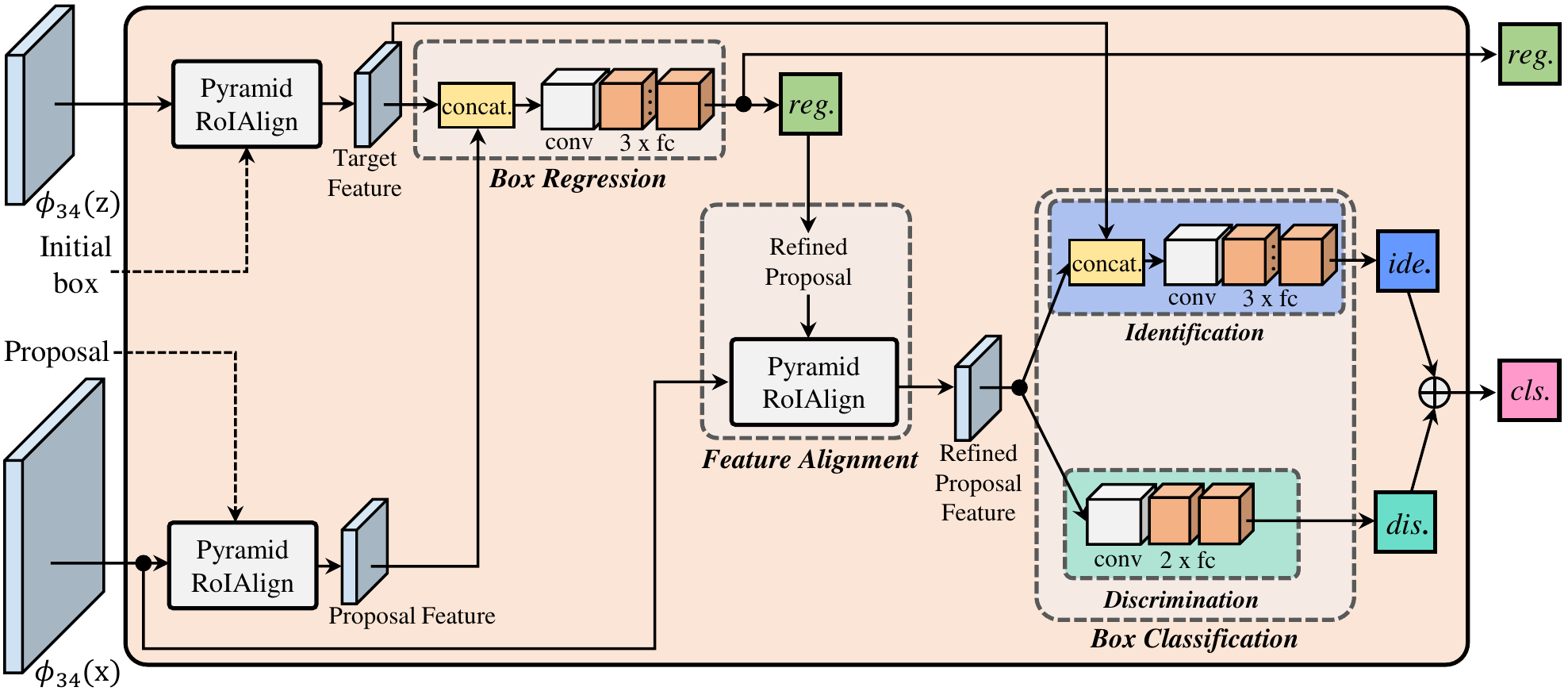}
	\caption{Illustration of CRAC module. \emph{Best viewed in color and by zooming-in}.}
	\label{fig:fig3}
\end{figure}

\subsection{CRAC for Proposal Refinement}
\label{RAC}

Because the proposals may contain distractors and/or not be good enough to handle large object scale variations, we develop a cascaded regression-align-classification (CRAC) module that refines each coarse proposal by cascading three steps, \ie, {\it box regression}, {\it feature alignment} and {\it box classification}, for better selection. Figure~\ref{fig:fig3} illustrates the architecture of CRAC. We show the detailed parameters of each component of CRAC in the supplementary material due to limited space.

\subsubsection{Box Regression}

Since only one-step regression of coarse proposals may not be sufficient to handle object scale changes, we employ an additional box regression in CRAC to further adjust locations and sizes of proposals. In specific, as shown in Figure~\ref{fig:fig3}, we first use pyramid RoIAlign (PRoIAligh) module (Section~\ref{proialign}) to extract the feature of each proposal. In order to improve regression accuracy, we employ features from multiple layers. Particularly, we concatenate the features $\phi_{4}(\mathbf{x})$ and $\phi_{3}(\mathbf{x})$ after blocks 4 and 3 and use a conv layer to obtain fused feature maps $\phi_{34}(\mathbf{x})$ (Figure~\ref{fig:fig2}). Afterwards, the feature $f_{{i}}$ of proposal $p_{i}$ is obtained through PRoIAlign as follows,
\begin{equation}
	f_{{i}} = \mathrm{PRoIAlign}(\phi_{34}(\mathbf{x}), p_{i})
\end{equation}

As a high-level task, we aim at learning a generic box regression model. Similar to the Siamese tracking~\cite{bertinetto2016fully,li2018high}, we incorporate the target in the first frame as prior information. Likewise, we use multi-level features and obtain the initial target feature as follows,
\begin{equation}
	\label{eq2}
	f_{\mathrm{init}} = \mathrm{PRoIAlign}(\phi_{34}(\mathbf{z}), b_{1})
\end{equation}
where $\phi_{34}(\mathbf{z})$  is fused feature maps for target (Figure~\ref{fig:fig2}) and $b_{1}$ denotes initial object box. Then, the box regression offset $r_{{i}}$ of $p_{i}$ is obtained via
\begin{equation}
	\label{offset}
	r_{{i}} = \mathcal{R}(f_{{i}}, f_{\mathrm{init}})
\end{equation}
where the box regression model $\mathcal{R}$ first concatenates $f_{{i}}$ and $f_{\mathrm{init}}$, and then applies a conv layer and three consecutive fc layers to output a 4-dimension vector $r_{{i}}=(r_{{i}}^{x}, r_{{i}}^{y}, r_{{i}}^{w}, r_{{i}}^{h})$. The loss $\ell_{\mathrm{reg}}$ to train the box regression model is smooth $L_1$ loss~\cite{girshick2015fast}.

\subsubsection{Feature Alignment}

\begin{figure}[!t]
	\centering
	\includegraphics[width=\linewidth]{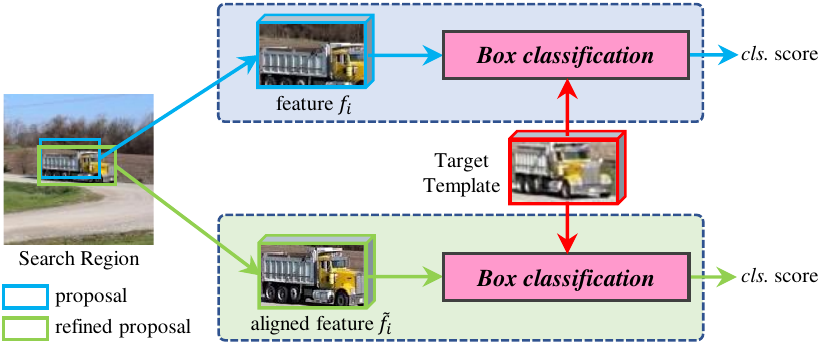}
	\caption{Comparison of proposal features with and without alignment. We observe that the aligned features are more accurate. \emph{Best viewed in color and by zooming in}.}
	\label{fig:fig4}
\end{figure}

Proposal classification is important, as it greatly affects final proposal selection. Existing refinement method (\eg,~\cite{wang2019spm}) directly extracts proposal features for classification. However, if the locations of proposals are inaccurate, their classification results may be degraded. Thanks to cascade structure of CRAC, we can alleviate this issue by aligning each proposal using offsets from box regression step. By doing so, more accurate proposal features can be used for classification. 

In particular, with regression offsets $r_{{i}}$ from Eq. (~\ref{offset}), we adjust location and size of $p_i$ as follows,
\begin{equation}
	\label{align}
	\begin{aligned}
		\tilde{x}_{i} &= x_{i} + w_{i}r_{{i}}^{x}   & \tilde{y}_{i} &= y_{i} + h_{i}r_{{i}}^{y}\\
		\tilde{w}_{i} &= w_{i}\mathrm{exp}(r_{{i}}^{w}) &\tilde{h}_{i} &= h_{i}\mathrm{exp}(r_{{i}}^{h})
	\end{aligned}
\end{equation}
where $x_{i}$, $y_{i}$, $w_{i}$, $h_{i}$ and $\tilde{x}_{i}$, $\tilde{y}_{i}$, $\tilde{w}_{i}$, $\tilde{h}_{i}$ represent the original and adjusted center coordinates of proposal $p_i$ and its width and height, respectively. With $\tilde{x}_{i}$, $\tilde{y}_{i}$, $\tilde{w}_{i}$, $\tilde{h}_{i}$, we can obtain the refined proposal $\tilde{p}_{i}$ for $p_{i}$, and extract more accurate feature using $\tilde{p}_{i}$ via
\begin{equation}
	\label{align}
	\tilde{f}_{{i}} = \mathrm{PRoIAlign}(\phi_{34}(\mathbf{x}), \tilde{p}_{i})
\end{equation}
where $\tilde{f}_{{i}}$ represents the aligned feature for $p_i$. In comparison with $f_{{i}}$, the aligned $\tilde{f}_{{i}}$ is more accurate (see Figure~\ref{fig:fig4}), which leads to better classification result. In addition, more accurate features can also benefit the training of box classification.

\subsubsection{Box Classification}

Since the proposals contain various distractors, a more discriminative classification module is desired in CRAC. Existing methods (\eg,~\cite{wang2019spm,fan2019siamese}) learn an additional matching sub-network to further classify the proposals for better selection. Owing to more balanced training samples, the classification model in refinement is more discriminative than that for proposal extraction. Despite this, these approaches still fail in presence of hard distractors due to ignorance of background information, which is crucial for distinguishing target from similar objects.

In this work, a joint {\it identification}-{\it discrimination} module is introduced in the box classification step of CRAC. Specifically, the identifier matches {\it offline} each proposal with reliable target template to find the most similar one. Different from the identifier, the discriminator learns {\it online} a classification model by exploiting background appearance information to suppress similar objects in proposals. By collaboration of these two components, our method enjoys both reliability of target template to select most similar proposal and the strong discriminative ability to suppress the difficult distractors, leading to robust classification.   

\vspace{0.5em}
\noindent
{\bf Identification.} The identifier aims to compute the similarities between proposals and target template. To this end, we leverage a relation network~\cite{sung2018learning} to 
learn offline a distance measurement between the template and a proposal owing to simplicity and efficiency, similar to~\cite{wang2019spm}. Since the identifier is learned to be generic, no update is required. As an advantage, the identifier will not be contaminated by background, and thus can resist accumulated errors in discrimination part caused by model update. We compute the identification score $\tilde{\nu}_{{i}}$ for refined proposal $\tilde{p}_{i}$ as follows,

\begin{equation}
	\label{ide}
	\tilde{\nu}_{{i}} = \mathcal{I}(\tilde{f}_{{i}}, f_{\mathrm{init}})
\end{equation}
where the identification model $\mathcal{I}$ first concatenates $\tilde{f}_{{i}}$ and $f_{\mathrm{init}}$, and then uses a conv layer and three fc layers to obtain a 2-dimension vector $\tilde{\nu}_{{i}}$, as shown in Figure~\ref{fig:fig3}. The loss $\ell_{\mathrm{ide}}$ to train the identification is cross entropy loss. 

\vspace{0.5em}
\noindent
{\bf Discrimination.} Different from the identifier, the discriminator focuses on suppressing similar distractors by exploiting background appearance information. For this purpose, we develop an online discrete-sampling-based classifier $\mathcal{D}$ with a light network architecture of one conv and two fc layers, as illustrated in Figure~\ref{fig:fig3}. We compute the discrimination score $\tilde{\tau}_{i}$ for $\tilde{p}_{i}$ as follows,
\begin{equation}
	\label{dis}
	\tilde{\tau}_{i}=\mathcal{D}(\tilde{f}_{i};\mathbf{w})
\end{equation}
where $\mathbf{w}$ denotes the parameters of the discrimination network.

To train discriminator, drawing inspiration from the success of discriminative regression tracking~\cite{danelljan2019atom,danelljan2017eco,lu2018deep,henriques2014high}, we use the $L_2$ loss to learn $\mathbf{w}$ as follows,

\begin{equation}
	\ell_{\mathrm{dis}} = \sum_{j=1}^{M}\|\mathcal{D}(X_j;\mathbf{w})-Y_j\|^{2} + \lambda\|\mathbf{w}\|^{2}
\end{equation}
where $X_j$ represents the feature of a training sample, $Y_j$ is a discrete (binary) label, and $\lambda$ is a regularization parameter. Notice that, unlike identifier trained on image pairs, we generate a set of discrete samples for training discriminator. We utilize the conjugate gradient method in~\cite{danelljan2019atom} to optimize discrimination network owing to its efficiency. We refer readers to~\cite{danelljan2019atom} for more details.

It is worth noting that, despite being relevant to discriminative regression tracking~\cite{danelljan2019atom,danelljan2017eco,lu2018deep,henriques2014high}, our discriminator is different in several aspects: (1) instead of performing classification on a large search region, our method only classifies a few discrete candidate proposals, which is more efficient; (2) the labels of training samples in our method are discrete (binary), which avoids boundary effects by using soft Gaussian labels as in~\cite{danelljan2019atom,danelljan2017eco,lu2018deep,henriques2014high}; and (3) because the training samples are discrete, we can easily implement the hard negative mining by focusing more on similar object regions in background.

With Eq. (\ref{ide}) and Eq. (\ref{dis}), we compute the box classification score $\tilde{s}_{i}$ for refined proposal $\tilde{p}_i$ via
\begin{equation}
	\label{cls}
	\tilde{s}_{i} = \alpha \cdot \tilde{\nu}_{i}^{+} + (1-\alpha) \cdot \tilde{\tau}_{i}
\end{equation}
where $\alpha$ is a trade-off parameter and $\tilde{\nu}_{i}^{+}$ denotes the positive classification score in $\tilde{\nu}_{i}$.

\subsection{Pyramid RoIAlign}
\label{proialign}
\begin{figure}
	\centering
	\includegraphics[width=0.9\linewidth]{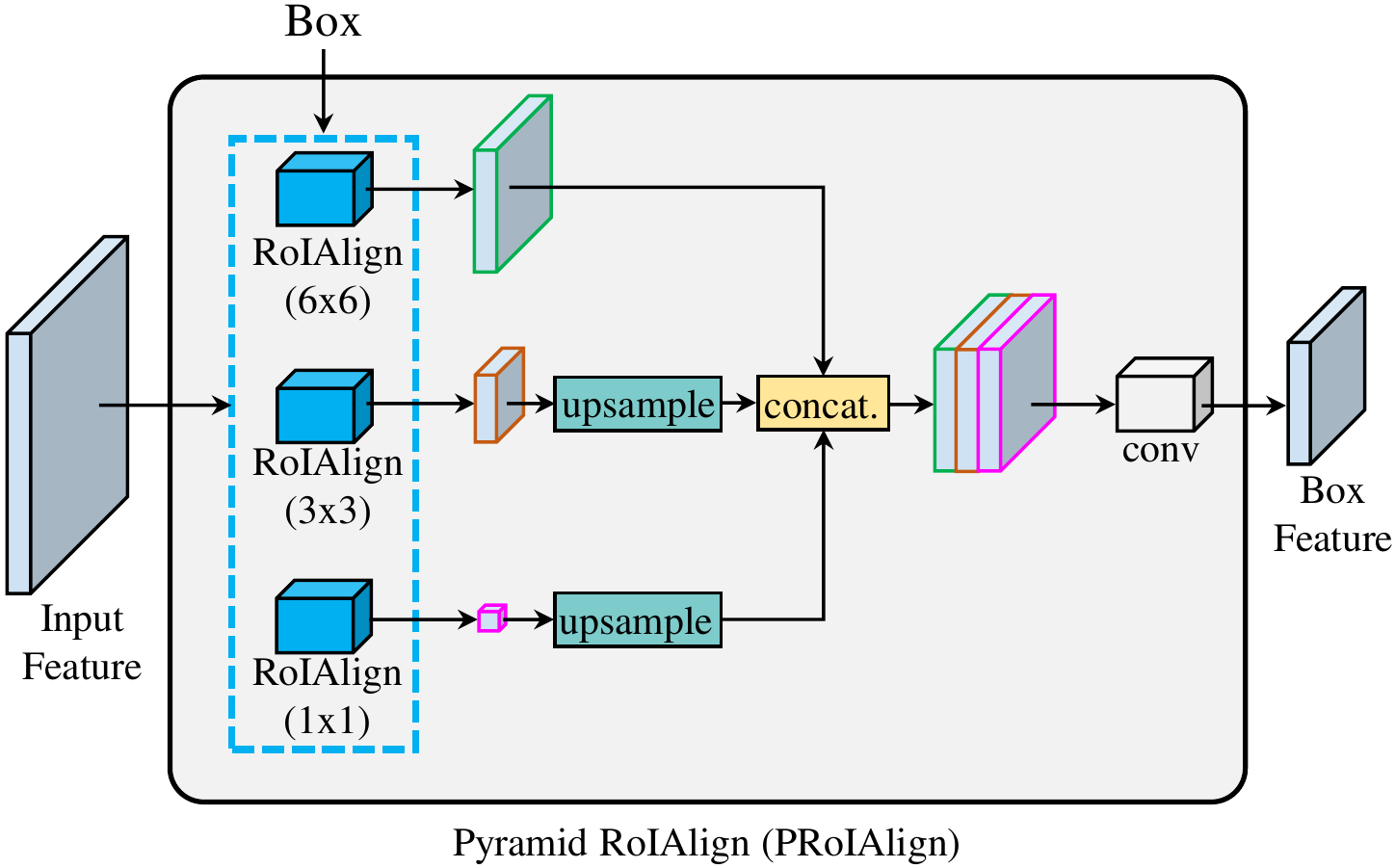}
	\caption{Illustration of pyramid RoIAlign.}
	\label{fig:fig5}
\end{figure}

Existing refinement approaches like~\cite{wang2019spm} adopt RoIAlign \cite{he2017mask} to extract proposal features. Specifically, the features of proposals are usually pooled to a fixed size (\eg, 6$\times$6). Despite simplicity, such features may be constrained to local target information and therefore sensitive to rotation and deformation. To alleviate this problem, we introduce a pyramid RoIAlign (PRoIAlign) module, which utilizes multiple RoIAlign operations to extract proposal features at different pooling sizes. For example, for size 1$\times$1, the proposal features contain global target information. To leverage both local and global cues, pooled features with different sizes are concatenated for fusion to derive more robust local-global proposal features. Figure~\ref{fig:fig5} illustrates the architecture of our PRoIAlign module. In our implementation, the PRoIAlign module is designed to have three levels, \ie, 6$\times$6, 3$\times$3 and 1$\times$1, for proposal feature extraction.

\begin{algorithm}[!t]\small
	\caption{Tracking with CRACT}
	\LinesNumbered 
	\KwIn{Image sequences $\{\mathbf{I}\}_{t=1}^{T}$, initial target box $b_{1}$ and trained model CRACT;}
	\KwOut{Tracking result $\{b_t\}_{t=2}^{T}$;}
	Crop target template $\bf{z}$ in $\bf{I}_{1}$ using $b_{1}$\;
	Extract feature embeddings $\phi_{34}(\bf{z})$ and $f_{\mathrm{init}}$ for $\bf{z}$\;
	\For{$t=2$ to $T$}{
		Crop the search region $\bf{x}$ in $\bf{I}_t$ using $b_{t-1}$\;
		Extract feature embedding $\phi_{34}(\bf{x})$ for $\bf{x}$\;
		Extract proposals $\{p_i\}_{i=1}^{N}$ $\leftarrow$ $\mathrm{RPN}(\phi_{34}(\bf{z}), \phi_{34}(\bf{x}))$\;
		Extract features $\{f_{i}\}_{i=1}^{N}$ for proposals\;
		Box regression to obtain $\{r_{i}\}_{i=1}^{N}$ using Eq. (\ref{offset})\;
		Feature alignment to obtain $\{\tilde{f}_{i}\}_{i=1}^{N}$ using Eq. (\ref{align})\;
		Box classification to obtain $\{\tilde{s}_{i}\}_{i=1}^{N}$ using Eq. (\ref{cls})\;
		Select the best proposal to determine the target box $b_{t}$ using Eq. (\ref{select}) \;
		Collect training samples based on $b_t$ and update the discriminator when necessary\;
	}
\end{algorithm}

\subsection{Training and Tracking}
\vspace{0.3em}
\noindent
{\bf Training.} The training of CRACT comprises two parts: (1) \textit{offline training} of Siamese RPN, box regression and identifier, and (2) \textit{online training} of discriminator in box classification. The first part is trained using image pairs, and the total training loss  $\mathcal{L}=\ell_{\mathrm{rpn}}+\ell_{\mathrm{reg}}+\ell_{\mathrm{ide}}$. Similar to~\cite{li2018high,wang2019spm}, the ratios of anchors are set to $[0.33,0.5,1,2,3]$ in RPN. The intersection over union (IoU) thresholds to determine anchors as positive (greater than threshold) or negative (less than threshold) are 0.6 and 0.3. We generate up to 64 samples from one image pair for RPN training. We choose at most 16 and 32 proposals for box regression and identifier training, respectively. The IoU thresholds to determine the proposals at positive (greater than threshold) or negative (less than threshold) are both 0.5. The second part is online trained during tracking. In particular, we draw 200 positive and 1000 negative samples in the first frame for initial training. The optimization strategy for training and update follows~\cite{danelljan2019atom} except training samples are discrete.

\vspace{0.3em}
\noindent
{\bf Tracking by Proposal Selection.} We formulate tracking as selecting the best proposal. For each sequence, we extract feature embeddings for target and initialize discriminator. When a new frame arrives, we crop a search region and perform RPN to generate proposals $\{p_{i}\}_{i=1}^{N}$, which are refined by CRAC to obtain $\{\tilde{p}_{i}\}_{i=1}^{N}$. We rank $\{\tilde{p}_{i}\}_{i=1}^{N}$ using coarse and refined classification scores and the target box $b$ is determined by the proposal with the highest score as follows,
\begin{equation}
	\label{select}
	b = \argmax_{\tilde{p}_{i}}(\beta \cdot \tilde{s}_{i} + (1-\beta) \cdot \tilde{c}_{i})
\end{equation}
where $\tilde{c}_{i}=c_{i}$ and $\tilde{s}_{i}$ denote respectively coarse and refined scores of $\tilde{p}_i$, and $\beta$ is a trade-off parameter. With tracking target box $b$, we collect $n^{+}$ positive and $n^{-}$ negative samples every $K$ frames to update the discriminator. We leverage short-long update strategy in~\cite{nam2016learning}. Notice that, we only update the two fc layers in the discrimination network. To improve robustness, we use hard negative mining by increasing the number of similar distractors in negative samples. Algorithm \textcolor{red}{1} summarizes the tracking with CRACT.

\section{Experiments}

\vspace{0.3em}
\noindent
{\bf Implementation.} We implement CRACT in python using PyTorch~\cite{paszke2019pytorch} on a single GTX 1080 GPU with 8GB memory. We utilize ResNet-18~\cite{he2016deep} as backbone and borrow its parameters trained on ImageNet~\cite{deng2009imagenet}. The number $N$ of proposals during tracking is empirically set to 10. The trade-off parameters $\alpha$ and $\beta$ are 0.4 and 0.8, respectively. The update interval $K$ for the discriminator is 10. $n^{+}$ and $n^{-}$ are set to 50 and 200, respectively. The learning rate of the offline training part is $10^{-2}$ with a decay of $10^{-4}$. It is trained end-to-end with SGD by 50 epochs. We apply LaSOT~\cite{fan2019lasot}, TrackingNet~\cite{muller2018trackingnet}, GOT-10k~\cite{huang2019got} and COCO~\cite{lin2014microsoft} for offline training, excluding the one under testing. The online training and update of the discriminator utilizes the strategy in~\cite{danelljan2019atom}. Hard negative mining is used in update. Our tracker runs at  28 frames per second ({\it fps}).

\begin{table}[!t]\scriptsize
	\centering
	\caption{Comparison with state-of-the-arts on OTB-2015~\cite{WuLY15}. The best three results highlighted in \BEST{red}, \SBEST{green} and \TBEST{blue}, respectively, throughout the rest of the paper.}
	\begin{tabular}{@{}rccc@{}}
		\toprule[1.5pt]
		Tracker & Where & PRE Score & SUC Score \\
		\hline\hline
		MDNet~\cite{nam2016learning} & CVPR'16     & 0.909 & 0.678 \\
		SiamFC~\cite{bertinetto2016fully} & ECCVW'16     & 0.771 & 0.582 \\
		ECO~\cite{danelljan2017eco} & CVPR'17     & 0.910 & 0.691 \\
		PTAV~\cite{fan2017parallel} & ICCV'17     & 0.849 & 0.635 \\
		SA-Siam~\cite{he2018twofold} & CVPR'18     & 0.865 & 0.657 \\
		DaSiamRPN~\cite{zhu2018distractor} & ECCV'18     & 0.880 & 0.658 \\
		SiamRPN++~\cite{li2019siamrpn++} & CVPR'19     & \TBEST{0.915} & 0.696 \\
		C-RPN~\cite{fan2019siamese} & CVPR'19     &0.885 & 0.663 \\
		SPM-18~\cite{wang2019spm}& CVPR'19     &0.912 & \TBEST{0.701} \\
		SiamDW~\cite{zhang2019deeper}& CVPR'19     &0.900 & 0.670 \\
		ATOM~\cite{danelljan2019atom} & CVPR'19     & 0.864 & 0.655 \\
		DiMP-50~\cite{bhat2019learning} & ICCV'19     & 0.900 & 0.688 \\
		SiamBAN~\cite{chen2020siamese}   & CVPR'20     & 0.910 & 0.696 \\
		Retina-MAML~\cite{wang2020tracking}   & CVPR'20     & n/a & \SBEST{0.712} \\
		SiamAttn~\cite{yu2020deformable} & CVPR'20     & \SBEST{0.926} & \SBEST{0.712} \\
		\hline
		CRACT (Ours)   & -     & \BEST{0.936} & \BEST{0.726} \\
		\toprule[1.5pt]
	\end{tabular}%
	\label{tab:otb}%
\end{table}%

\begin{table*}[!t]\scriptsize
	\centering
	\caption{Comparison with state-of-the-art trackers on UAV123~\cite{mueller2016benchmark}. 
	}
	\begin{tabular}{@{}R{1.1cm}C{0.85cm}C{0.85cm}C{0.85cm}C{0.85cm}C{1cm}C{0.8cm}C{1cm}C{0.85cm}C{0.95cm}C{0.95cm}C{0.95cm}C{0.9cm}@{}}
		\toprule[1.5pt]
		Tracker & \tabincell{c}{ECOhc\\\cite{danelljan2017eco}} & \tabincell{c}{ECO\\\cite{danelljan2017eco}}   & \tabincell{c}{SiamRPN\\\cite{li2018high}} & \tabincell{c}{RT-MDNet\\\cite{jung2018real}}  & \tabincell{c}{DaSiam\\RPN~\cite{zhu2018distractor}} & \tabincell{c}{ARCF\\\cite{huang2019learning}}  & \tabincell{c}{SiamRPN\\++~\cite{li2019siamrpn++}} & \tabincell{c}{ATOM\\\cite{danelljan2019atom}}  & \tabincell{c}{DiMP-50\\\cite{bhat2019learning}} & \tabincell{c}{SiamBAN\\\cite{chen2020siamese}} & \tabincell{c}{SiamAttn\\\cite{yu2020deformable}} & \tabincell{c}{CRACT\\(ours)} \\
		\hline\hline
		Where & CVPR'17 & CVPR'17 & CVPR'18 & ECCV'18 & ECCV'18 & CVPR'19 & CVPR'19 & CVPR'19 & ICCV'19 & CVPR'20 & CVPR'20 & - \\
		PRE  & 0.725 & 0.741 & 0.748 & 0.772   & 0.796 & 0.670 & 0.807 & \TBEST{0.856} & \SBEST{0.858} & 0.833 & 0.845 & \BEST{0.860} \\
		SUC  & 0.506 & 0.525 & 0.527 & 0.528 & 0.586 & 0.470 & 0.613 & 0.642 & \SBEST{0.653} & 0.631 & \TBEST{0.650} & \BEST{0.664} \\
		\toprule[1.5pt]
	\end{tabular}%
	\label{tab:uav123}%
\end{table*}%

\begin{table*}[!t]\scriptsize
	\centering
	\caption{Comparison with state-of-the-art trackers on NfS~\cite{kiani2017need}. 
	}
	\begin{tabular}{@{}R{1.1cm}C{0.85cm}C{0.85cm}C{0.85cm}C{0.85cm}C{1.2cm}C{0.85cm}C{0.85cm}C{0.85cm}C{0.95cm}C{0.95cm}C{0.95cm}C{0.9cm}@{}}
		\toprule[1.5pt]
		Tracker & \tabincell{c}{HCF\\\cite{ma2015hierarchical}}   & \tabincell{c}{HDT\\\cite{qi2016hedged}}   & \tabincell{c}{MDNet\\\cite{nam2016learning}} & \tabincell{c}{SiamFC\\\cite{bertinetto2016fully}} & \tabincell{c}{ECOhc\\\cite{danelljan2017eco}} & \tabincell{c}{ECO\\\cite{danelljan2017eco}}   & \tabincell{c}{BACF\\\cite{kiani2017learning}}  & \tabincell{c}{UPDT\\\cite{bhat2018unveiling}}  & \tabincell{c}{ATOM\\\cite{danelljan2019atom}}  & \tabincell{c}{DiMP-50\\\cite{bhat2019learning}} & \tabincell{c}{SiamBAN\\\cite{chen2020siamese}} & \tabincell{c}{CRACT\\(ours)} \\
		\hline\hline
		Where & ICCV'15 & CVPR'16 & CVPR'16 & ECCVW'16 & CVPR'17 & CVPR'17 & ICCV'17 & ECCV'18 & CVPR'19 & ICCV'19 & CVPR'20 &  - \\
		SUC  & 0.295 & 0.403 & 0.429 & 0.401 & 0.459 & 0.466 & 0.341 & 0.542 & 0.590 & \SBEST{0.619} & \TBEST{0.594} & \BEST{0.625} \\
		\toprule[1.5pt]
	\end{tabular}%
	\label{tab:nfs}%
\end{table*}%

\begin{table*}[!t]\scriptsize
	\centering
	\caption{Comparison with other trackers on VOT-2018~\cite{kristan2018sixth}. 
	}
	\begin{tabular}{@{}R{0.85cm}C{1cm}C{0.85cm}C{0.85cm}C{0.85cm}C{1cm}C{0.85cm}C{0.85cm}C{0.85cm}C{0.95cm}C{0.95cm}C{1cm}C{0.9cm}@{}}
		\toprule[1.5pt]
		Tracker & \tabincell{c}{SiamFC\\\cite{bertinetto2016fully}} & \tabincell{c}{ECO\\\cite{danelljan2017eco}}   & \tabincell{c}{SA-Siam\\\cite{he2018twofold}} & \tabincell{c}{SiamRPN\\\cite{li2018high}} & \tabincell{c}{UPDT\\\cite{bhat2018unveiling}}  & \tabincell{c}{DaSiam\\RPN~\cite{zhu2018distractor}} & \tabincell{c}{SiamRPN\\++~\cite{li2019siamrpn++}} & \tabincell{c}{ATOM\\\cite{danelljan2019atom}}  & \tabincell{c}{DiMP-50\\\cite{bhat2019learning}} & \tabincell{c}{SiamBAN\\\cite{chen2020siamese}} & \tabincell{c}{Retina-\\MAML~\cite{wang2020tracking}} & \tabincell{c}{CRACT\\(ours)} \\
		\hline\hline
		Where & ECCVW'16 & CVPR'17 & CVPR'18 & CVPR'18 & ECCV'18 & ECCV'18 & CVPR'19 & CVPR'19 & ICCV'19 & CVPR'20 & CVPR'20 & - \\
		Acc.  & 0.500 & 0.480 & 0.543 & 0.588 & 0.536 & 0.590 & \TBEST{0.600} & 0.590 & 0.597 & 0.597 & \SBEST{0.604} & \BEST{0.611} \\
		Rob.  & 0.590 & 0.280 & 0.224 & 0.276 & 0.184 & 0.280 & 0.234 & 0.204 & \BEST{0.153} & 0.178 & \SBEST{0.159} & \TBEST{0.175} \\
		EAO   & 0.188 & 0.276 & 0.325 & 0.384 & 0.376 & 0.383 & 0.414 & 0.401 & \TBEST{0.440} & \SBEST{0.452} & \SBEST{0.452} & \BEST{0.455} \\
		\toprule[1.5pt]
	\end{tabular}%
	\label{tab:vot18}%
\end{table*}%

\subsection{State-of-the-art Comparison}

\vspace{0.3em}
\noindent
{\bf OTB-2015~\cite{WuLY15}.} OTB-2015 is a popular tracking benchmark with 100 videos. We compare CRACT with 15 trackers. The comparison is demonstrated in Table~\ref{tab:otb} with precision (PRE) and success (SUC) scores using one-pass evaluation (OPE). CRACT achieves the best results with 0.936 PRE score and 0.726 SUC score, outperforming the second best by 1.0\% and 1.4\%, respectively. Compared with SiamRPN++ with 0.915 PRE score and 0.696 SUC score, we achieve 2.1\% and 3.0\% gains owing to RAC. Besides, compared to proposal refinement method SPM-18, which can serve as our baseline, with 0.912 PRE score and 0.701 SUC score, CRACT with cascaded refinement shows 2.4\% and 2.5\% improvements, evidencing effectiveness in boosting tracking robustness and accuracy.

\vspace{0.3em}
\noindent
{\bf UAV123~\cite{mueller2016benchmark}.} UAV123 focuses on aerial object tracking and contains 123 videos. We compare CRACT to 11 trackers and the results are displayed in Table~\ref{tab:uav123}. CRACT obtains the best 0.860 PRE score and 0.664 SUC score, outperforming the second best DiMP-50 with 0.858 PRE score and 0.653 SUC score. In comparison to SiamRPN++ with 0.613 SUC score, we achieve 5.1\% absolute gain, which clearly shows the advantage of our proposal refinement. Moreover, CRACT also outperforms the recent anchor-free SiamBAN by 3.3\% in term of SUC score.

\begin{table}[!t]\scriptsize
	\centering
	\caption{Comparison with other trackers on TrackingNet~\cite{muller2018trackingnet}. 
	}
	\begin{tabular}{@{}rC{0.8cm}C{1.08cm}cC{1.1cm}@{}}
		\toprule[1.5pt]
		Tracker & Where & PRE Score  & NPRE Score  & SUC Score \\
		\hline\hline
		C-RPN~\cite{fan2019siamese} & CVPR'19 & 0.619 & 0.746 & 0.669 \\
		SiamRPN++~\cite{li2019siamrpn++} & CVPR'19 & 0.694 & 0.799 & 0.733 \\
		SPM~\cite{wang2019spm}   & CVPR'19 & 0.661 & 0.778 & 0.712 \\
		ATOM~\cite{danelljan2019atom}  & CVPR'19 & 0.648 & 0.771 & 0.703 \\
		DiMP-50~\cite{bhat2019learning} & ICCV'19 & n/a     & \TBEST{0.801} & \TBEST{0.740} \\
		Retina-MAML~\cite{wang2020tracking} & CVPR'20 & n/a     & 0.786 & 0.698 \\
		SiamAttn~\cite{yu2020deformable} & CVPR'20 & n/a     & \SBEST{0.817} & \SBEST{0.752} \\
		\hline
		CRACT (ours) & -     & \BEST{0.724} & \BEST{0.824} & \BEST{0.754} \\
		\toprule[1.5pt]
	\end{tabular}%
	\label{tab:trackingnet}%
\end{table}%

\begin{figure}[!t]
	\centering
	\includegraphics[width=\linewidth]{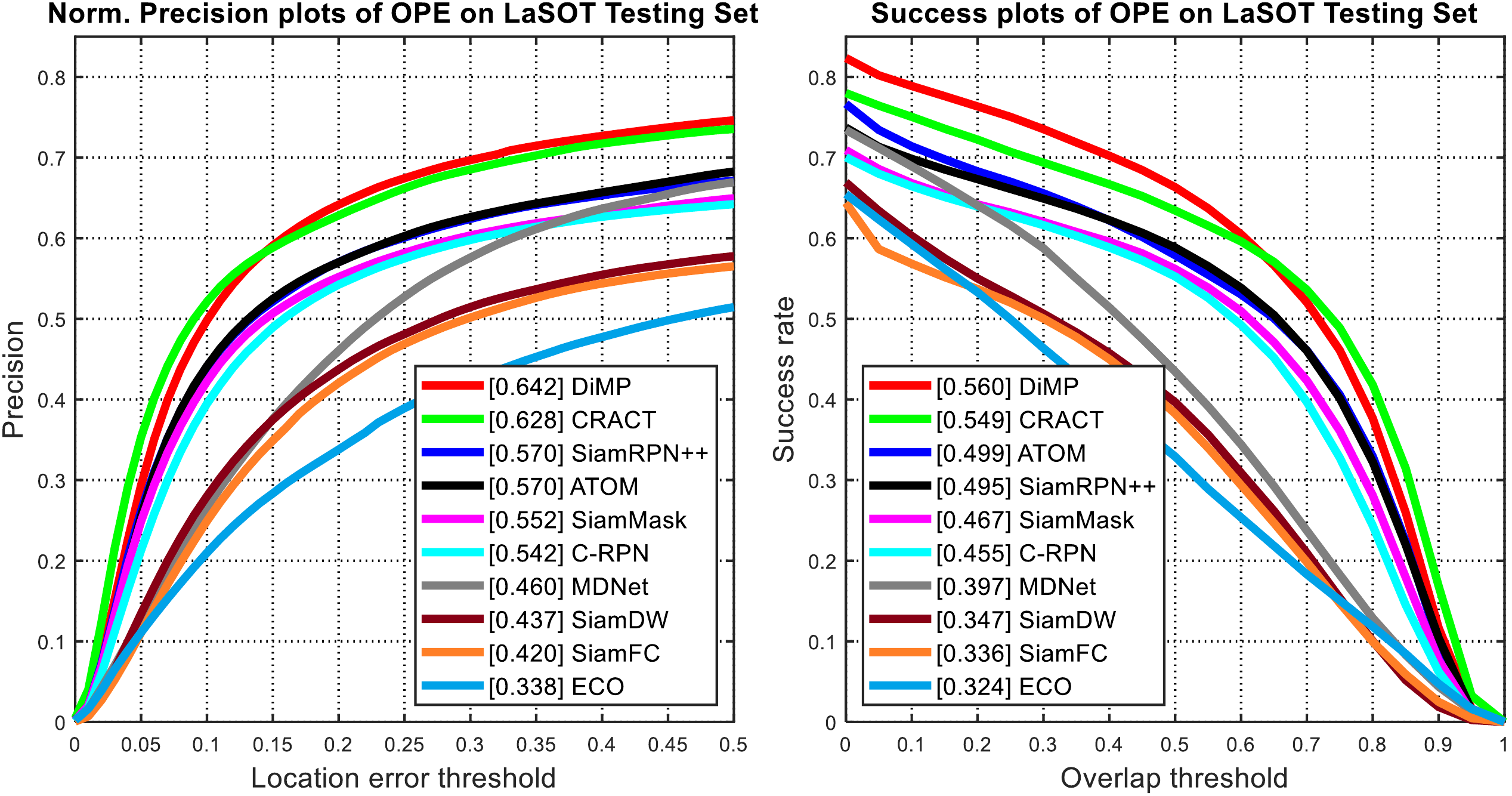}
	\caption{Comparison with state-of-the-arts on LaSOT~\cite{fan2019lasot}. \emph{Best viewed in color and by zooming in}.}
	\label{fig:lasot}
\end{figure}

\vspace{0.3em}
\noindent
{\bf NfS~\cite{kiani2017need}.} NfS consists of 100 sequences for evaluation on high frame rate videos. We evaluate our approach on 30 {\it fps} version. Table~\ref{tab:nfs} demonstrates our result and comparison to 11 trackers. Our CRACT achieves the best result with 0.625 SUC score, which outperforms the second best DiMP-50 with 0.619 SUC score by 0.6\% and the third best SiamBAN with 0.594 by 3.1\%.

\vspace{0.3em}
\noindent
{\bf VOT-2018~\cite{kristan2018sixth}.} VOT-2018 contains 60 videos for tracking. We compare CRACT with 11 trackers and Table~\ref{tab:vot18} demonstrates the comparison results. Our tracker achieves the best of 0.455 on EAO. Compared to SiamRPN++ which also regards tracking as proposal selection, CRACT obtains a performance gain of 4.1\% in term of EAO, which shows the effectiveness of our hierarchical RAC in refining proposals for better selection. Compared to the recent state-of-the-art DiMP-50 with 0.440 EAO score, our method achieves 1.5\% improvement. Moreover, CRACT outperforms SiamBAN and Retina-MAML both with 0.452 EAO score.

\begin{table}[!t]\scriptsize
	\centering
	\caption{Comparison results on GOT-10k~\cite{huang2019got}. 
	}
	\begin{tabular}{@{}C{0.8cm}C{0.8cm}C{1cm}C{0.7cm}C{0.7cm}C{0.8cm}C{0.9cm}@{}}
		\toprule[1.5pt]
		Tracker & \tabincell{c}{MDNet\\\cite{nam2016learning}} & \tabincell{c}{SiamFC\\\cite{bertinetto2016fully}} & \tabincell{c}{SPM\\\cite{wang2019spm}}   & \tabincell{c}{ATOM\\\cite{danelljan2019atom}}  & \tabincell{c}{DiMP-50\\\cite{bhat2019learning}} & \tabincell{c}{CRACT\\(ours)} \\
		\hline\hline
		Where & CVPR'16 & ECCVW'16 & CVPR'19 & CVPR'19 & ICCV'19 & - \\
		AO    & 0.299 & 0.348 & 0.513 & \TBEST{0.556} & \SBEST{0.611} & \BEST{0.620} \\
		SR$_{0.50}$ & 0.303 & 0.353 & 0.593 & \TBEST{0.634} & \SBEST{0.717} & \BEST{0.728} \\
		SR$_{0.75}$ & 0.099 & 0.098 & 0.359 & \TBEST{0.402}& \SBEST{0.492} & \BEST{0.496} \\
		\toprule[1.5pt]
	\end{tabular}%
	\label{tab:got}%
\end{table}%

\vspace{0.3em}
\noindent
{\bf TrackingNet~\cite{muller2018trackingnet}.} TrackingNet offers 511 videos for evaluation. Table~\ref{tab:trackingnet} shows comparison results of CRACT with 7 state-of-the-art trackers. Our method achieves the best results of 0.724, 0.824 and 0.754 on PRE, NPRE and SUC scores, outperforming recent trackers SiamAttn and DiMP-50. In addition, compared to SiamRPN++ with 0.733 SUC score and SPM with 0.712 SUC score, we obtain performance gains of 2.1\% and 4.2\%, respectively, evidencing the advantage of our hierarchical refinement.

\vspace{0.3em}
\noindent
{\bf LaSOT~\cite{fan2019lasot}.} LaSOT is a recent long-term tracking benchmark. We evaluate our approach under protocol \uppercase\expandafter{\romannumeral2} in which 280 videos are provided for testing. Figure~\ref{fig:lasot} shows our results and comparison with 9 state-of-the-arts. CRACT achieves the second best results with 0.628 normalized PRE score and 0.549 SUC score, slightly lower than the 0.642 normalized PRE score and 0.560 SUC score by DiMP-50. Compared with ATOM and SiamRPN++ with 0.499 and 0.495 SUC scores, CRACT shows clear performance gains of 5.0\% and 5.4\%.

\vspace{0.3em}
\noindent
{\bf GOT-10k~\cite{huang2019got}.} GOT-10k offers 180 challenging videos for short-term tracking evaluation. We compare CRACT to 5 trackers as displayed in Table~\ref{tab:got}. CRACT performs the best with 0.620 AO score, outperforming the second best DiMP-50 with 0.611 AO score. Besides, CRACT obtains a significant performance gain of 10.7\% compared to SPM.

Due to limited space, we demonstrate qualitative tracking results and comparisons in supplementary material.

\subsection{Ablation Study}

To verify each component in CRACT, we conduct ablative experiments on OTB-2015~\cite{WuLY15} and NfS~\cite{kiani2017need}.

\vspace{0.3em}
\noindent
{\bf Cascade structure.} In this paper, we introduce a novel proposal refinement module with cascade structure. We verify its effectiveness by designing a refinement module with parallel structure by removing feature alignment (see detailed architecture in supplementary material). Table~\ref{tab:hierarchical} shows results of parallel and cascade refinement. We observe that CRACT with parallel refinement achieves SUC scores of 0.713 and 0.609 on OTB-2015 and NfS. By utilizing cascaded proposal refinement, the results are significantly improved to 0.726 (1.3\% gain) and 0.625 (1.6\% gain), which clearly evidences the advantage of using more accurately regressed proposals for proposal selection.

\vspace{0.3em}
\noindent
{\bf Identification-discrimination.} We propose a joint module of discrimination and discrimination in CRAC for proposal classification. In fact, either the identifier or discriminator can be used individually for proposal classification. However, each has advantages and disadvantages. The identifier can easily recognize the target from non-semantic distractors using powerful distance measurement. In addition, it avoids the contamination by background owing to no update. Nevertheless, it cannot leverage appearance information. The discriminator works well in suppressing semantic distractors through online learning background information. Nonetheless, it has a risk of model contamination caused by update. By collaboration of identifier and discriminator, they can complement each other for better robust proposal selection. We verify the effects of individual and joint use of identifier and discriminator. Table~\ref{tab:ide-dis} shows the comparison. Using identifier only and discriminator only achieves SUC scores of 0.715 and 0.712 on OTB-215. with joint consideration of them, the performance is significantly boosted to 0.726. Likewise, the best result of 0.625 SUC score is obtained when combining identifier and discriminator. 

\begin{table}[!t]\small
	\centering
	\caption{Comparison of simultaneous and hierarchical refinement.}
	\begin{tabular}{@{\hspace{1.2mm}}r@{\hspace{1.2mm}}cc@{\hspace{1.2mm}}}
		\toprule[1.5pt]
		& \tabincell{c}{Parallel refinement} & \tabincell{c}{Cascaded refinement} \\
		\hline\hline
		SUC on OTB-2015 & 0.713 & 0.726 \\
		\hline
		SUC on NfS & 0.609 & 0.625 \\
		\toprule[1.5pt]
	\end{tabular}%
	\label{tab:hierarchical}%
\end{table}%

\begin{table}[!t]\small
	\centering
	\caption{Comparison (in SUC) between individual and joint use of identifier and discriminator.}
	\begin{tabular}{@{\hspace{2mm}}rccc@{\hspace{2mm}}}
		\toprule[1.5pt]
		& Identifier only &  Discriminator only & Joint \\
		\hline\hline
		OTB-2015 & 0.715 & 0.712 & 0.726 \\
		\hline
		NfS & 0.606 & 0.614 & 0.625 \\
		\toprule[1.5pt]
	\end{tabular}%
	\label{tab:ide-dis}%
\end{table}%

\begin{table}[!t]\small
	\centering
	\caption{Comparison between RoIAlign and pyramid RoIAlign.}
	\begin{tabular}{@{}rcc@{}}
		\toprule[1.5pt]
		& RoIAlign & PRoIAlign \\
		\hline\hline
		SUC on OTB-2015 & 0.719 & 0.726 \\
		\hline
		SUC on NfS & 0.615 & 0.625 \\
		\toprule[1.5pt]
	\end{tabular}%
	\label{tab:roi}%
\end{table}%

\vspace{0.3em}
\noindent
{\bf Pyramid RoIAlign.} Different from current tracker~\cite{wang2019spm} using
RoIAlign~\cite{he2017mask} for proposal extraction, we present a simple yet effective PRoIAlign to exploit global and local cues. Table~\ref{tab:roi} shows the results with RoIAlign and our PRoIAlign. We observe that PRoIAlign improves the SUC scores from 0.719 to 0.716 on OTB-2015 and from 0.615 to 0.625 on NfS, respectively, showing the advantage of exploring various cues in performance improvement.

\section{Conclusion}

In this paper, we propose a novel tracker dubbed CRACT for accurate and robust tracking. CRACT first extracts a few coarse proposals and then refines each proposal using the proposed cascaded regression-align-classification module. During inference, the best proposal determined by both coarse and refined classification scores is selected to be the final target. Experiments on seven benchmarks demonstrate its superior performance. In the future, we plan to improve the performance of CRACT by integrating mask segmentation into our cascade refinement.

{\small
\bibliographystyle{ieee_fullname}
\bibliography{egbib}
}

\end{document}